\documentclass[lettersize,journal]{IEEEtran}

\usepackage{amsmath,amsfonts}
\usepackage{algorithmic}
\usepackage{algorithm}
\usepackage{array}
\usepackage[caption=false,font=footnotesize,labelfont=sf,textfont=sf]{subfig}
\usepackage{textcomp}
\usepackage{stfloats}
\usepackage{url}
\usepackage{verbatim}
\usepackage{graphicx}
\usepackage{cite}
\usepackage{hyperref}
\usepackage{orcidlink}
\usepackage{booktabs}
\usepackage{multirow}
\usepackage{balance}

\hypersetup{
    colorlinks=true,
    linkcolor=blue,
    urlcolor=blue,
    citecolor=blue,
}

\makeatletter
\def\section{\@startsection{section}{1}{\z@}{1.5ex \@plus 0.5ex \@minus 0.5ex}%
{0.5ex \@plus 0.5ex \@minus 0ex}{\normalfont\normalsize\centering\scshape}}
\def\subsection{\@startsection{subsection}{2}{\z@}{1.0ex \@plus 0.5ex \@minus 0.5ex}%
{0.5ex \@plus 0.5ex \@minus 0ex}{\normalfont\normalsize\itshape}}
\makeatother

\begin{document}

\title{Computational Depth Measurement in Thermographic Video: Overcoming Spatial Overfitting via Spatio-Temporal Decoupling}

\author{Zain~Ul~Abidin~\orcidlink{0009-0008-2708-0385}, Habeeban~Memon~\orcidlink{0009-0002-0276-4322}, and~Junaid~Ahmed~\orcidlink{0000-0003-0083-8068}%
\thanks{Zain Ul Abidin (Corresponding Author), Habeeban Memon, and Junaid Ahmed are with the Department of Computer Systems Engineering, Sukkur IBA University, Sukkur, Pakistan (e-mail: 
\href{mailto:zainulabdin995@gmail.com}{zainulabdin995@gmail.com};
\href{mailto:habeebanmemon73@gmail.com}{habeebanmemon73@gmail.com};  \href{mailto:j.bhatti@iba-suk.edu.pk}{j.bhatti@iba-suk.edu.pk}).}}

\markboth{}%
{Memon \MakeLowercase{\textit{et al.}}: Computational Measurement of CFRP Subsurface Defects}

\maketitle

\begin{abstract}
Accurate through-thickness measurement of subsurface delamination depth in Carbon Fiber Reinforced Polymer (CFRP) is important for structural assessment because defect location determines affected load-bearing layers. Optical pulsed thermography (OPT) provides a two-dimensional thermal video rather than volumetric measurements, so depth must be inferred from temporal heat-diffusion responses. A challenge is spatial dataset bias: when calibration defects follow regular grids, regression models may memorize their geometry instead of learning physical relationship between thermal decay and depth. This work introduces a spatio-temporal decoupling architecture that separates spatial defect localization from temporal depth measurement. Defect regions are first localized using segmentation methods, after which thermal responses are spatially averaged and converted into sixteen physics-informed temporal, energy, statistical, and geometric features. These features expose the one-dimensional heat-conduction relationship while withholding pixel coordinates from the depth model. Four regression models are evaluated using specimen-level cross-validation: Random Forest (RF), Gradient Boosting Machine (GBM), Advanced Multi-Layer Perceptron (Adv-MLP), and XGBoost. Unregularized trees and over-parameterized Adv-MLP exhibit calibration collapse under geometric shifts, with errors exceeding 0.5 mm. In contrast, regularized XGBoost with L1/L2 penalties and column sampling maintains cross-specimen calibration, achieving a mean absolute error (MAE) of 0.056 mm and root mean square error (RMSE) of 0.085 mm. Predicted depths are merged with masks to generate Delaunay-triangulated three-dimensional defect models in three to five seconds per specimen. Results show that mathematical regularization and spatio-temporal decoupling reduce spatial memorization in thermal-video depth regression.
\end{abstract}


% ==========================================
% SECTION I: INTRODUCTION
% ==========================================
\section{Introduction}
\label{sec:introduction}
\IEEEPARstart{I}{n} safety-critical aerospace and renewable-energy systems, Carbon Fiber Reinforced Polymer (CFRP) composites are widely used because of their high strength-to-weight ratio \cite{ref1}. Interlaminar delamination is a critical failure mode and can originate from low-energy impacts such as tool drops or bird strikes that leave little or no visible surface damage \cite{ref2}. The through-thickness position of a hidden defect is particularly important because defects at different depths affect different laminate layers and therefore different load-transfer paths \cite{ref3}. Consequently, binary defect detection alone is insufficient for structural assessment; a useful inspection system must provide an absolute or calibrated estimate of defect depth.

Optical pulsed thermography (OPT) is a noncontact inspection technique in which a short optical excitation heats the specimen surface while an infrared camera records the transient cooling process \cite{ref4,ref5}. The measured data are naturally represented as a thermal tensor $X(i,j,t)$, where $(i,j)$ denotes image coordinates and $t$ denotes the acquisition frame. A subsurface discontinuity changes the local heat-flow path and consequently changes the surface thermal response. Under a one-dimensional heat-conduction approximation, defect depth $d$ and peak contrast time $t_{\mathrm{peak}}$ are related by \cite{ref6}
\begin{equation}
\label{eq:heat_conduction}
d=\beta\sqrt{\alpha t_{\mathrm{peak}}},
\end{equation}
where $\alpha$ is thermal diffusivity and $\beta$ is a geometric factor. Equation \eqref{eq:heat_conduction} provides the physical basis for treating temporal response as a measurement variable rather than treating image position as a depth label.

Classical thermographic processing has established several ways to expose this temporal information. Thermographic signal reconstruction can estimate characteristic thermal times while suppressing noise \cite{ref7,ref8}; pulsed phase thermography reduces sensitivity to nonuniform heating \cite{ref9}; and principal component thermography can improve defect contrast and localization \cite{ref10}. Analytical depth estimation, however, remains sensitive to specimen-dependent thermal properties and calibration conditions \cite{ref11}. Variations in thermal diffusivity, geometry, and heating conditions can therefore limit direct application of \eqref{eq:heat_conduction} without an appropriate reference calibration.

Supervised learning has consequently become attractive for automated thermographic inspection. Deep-learning approaches have demonstrated automated localization of CFRP damage \cite{ref12,ref13}, while physics-informed feature extraction and thermal-signal learning have been investigated for quantitative inspection \cite{ref14}--\cite{ref20}. Other studies have considered one-dimensional thermal-signal learning, direct sequenced-signal depth estimation, and joint detection/depth architectures \cite{ref21}--\cite{ref24}. A three-dimensional convolutional approach can process the complete thermal volume, but its computational burden grows with the number of pixels and frames. For an input of spatial dimensions $H\times W$ and temporal length $T$, full volumetric processing scales with $\mathcal{O}(HWT)$, which becomes increasingly expensive as spatial resolution and sequence length increase \cite{ref15,ref25,ref26}.

Beyond computational cost, however, full spatial processing creates a more fundamental computer-vision problem: dataset bias. Laboratory depth datasets frequently use discrete flat-bottom holes or Teflon inserts arranged in highly regular layouts. In such data, spatial position is correlated with the target depth. A model can therefore obtain apparently excellent test accuracy without learning the underlying thermal physics if the train and test specimens share the same geometric arrangement. An unregularized tree ensemble can repeatedly partition feature space according to patterns correlated with the calibration grid, while an over-parameterized neural network can memorize the same structure. The resulting model is accurate only while the hidden spatial proxy remains available.

For thermal-video regression, simply supplying the full spatial tensor is not a cure: it increases access to the geometric coordinates responsible for the bias. The central issue is therefore representation invariance, not model capacity alone.

We address this issue through a spatio-temporal decoupling architecture. Spatial processing is restricted to defect localization. After localization, the pixels belonging to each defect are spatially averaged into a one-dimensional temporal thermal response. The depth model therefore receives temporal and derived physical descriptors instead of the original image coordinates. The spatial information is retained only for the final reconstruction, where the predicted depth is mapped back onto the detected defect boundary.

The proposed framework makes five contributions:
\begin{enumerate}
\item It formulates thermographic depth regression as a dataset-bias problem in which spatial geometry can act as a hidden depth label.
\item It separates spatial localization from temporal measurement through a two-stage spatio-temporal decoupling architecture, avoiding direct full-tensor regression.
\item It uses sixteen physics-informed features as explicit mathematical constraints derived from the thermal response and the one-dimensional heat-conduction relationship.
\item It provides a controlled diagnostic comparison of RF, GBM, Adv-MLP, and XGBoost under strict specimen-level cross-validation, with Fold 2 serving as the principal geometric distribution-shift test.
\item It converts the predicted depth and segmentation masks into three-dimensional Delaunay-triangulated defect models in three to five seconds per specimen.
\end{enumerate}

A direct numerical comparison with published deep-learning depth-estimation systems is not used as the primary benchmark because reported systems differ in hardware, flash-energy conditions, sampling frequencies, datasets, and experimental protocols. Instead, all four regression algorithms are evaluated under identical specimen partitions and identical feature representations. This controlled comparison isolates the effect of model regularization and representation on cross-geometry generalization.

% ==========================================
% SECTION II: METHODOLOGY
% ==========================================
\section{Spatio-Temporal Decoupling Measurement Methodology}
\label{sec:methodology}

\begin{figure*}[!tb]
\centering
\includegraphics[width=6.5in]{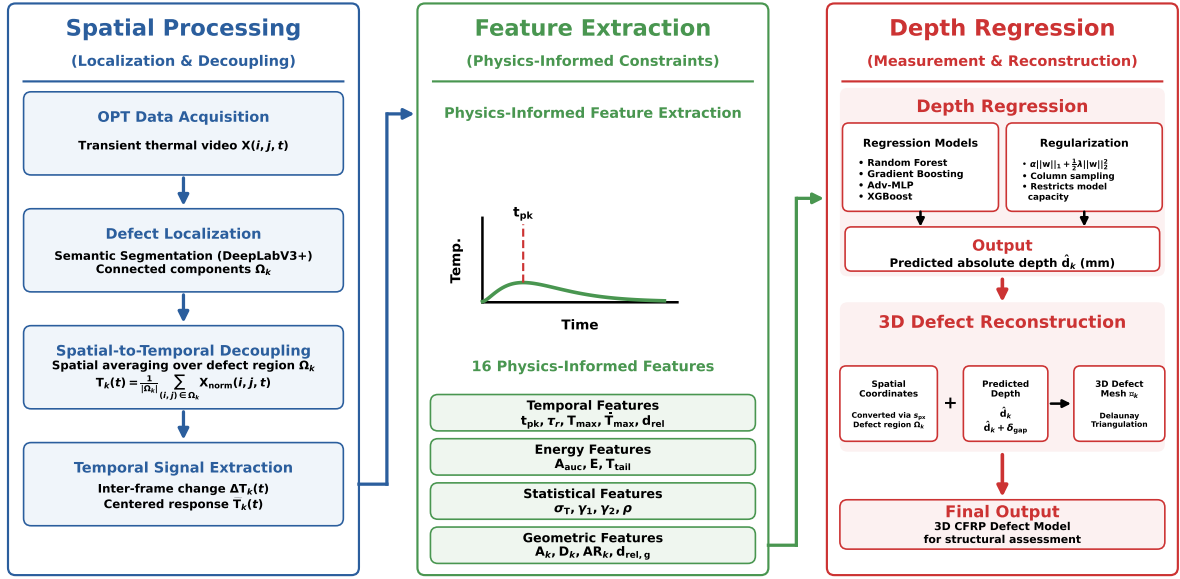}
\caption{Spatio-temporal decoupling architecture for depth estimation from thermographic video. The pipeline separates spatial localization from temporal measurement to avoid geometric memorization.}
\label{fig_pipeline}
\end{figure*}

The overall workflow of the spatio-temporal decoupling measurement architecture is outlined in Fig.~\ref{fig_pipeline}, illustrating the structured separation between spatial defect boundary extraction and temporal regression modeling.

\subsection{Stage 1: Spatial Defect Localization}

The first stage identifies the spatial support of each subsurface defect. The localization component is not proposed as a new segmentation architecture. Instead, it utilizes a pretrained DeepLabV3+ model acting on a cyclically aligned three-channel thermographic fusion. As detailed in \cite{ref27}, this input representation integrates raw thermal intensity, principal component thermography (PCT) spatial components, and thermographic signal reconstruction (TSR) temporal coefficient images. Because this representation effectively isolates defect boundaries from background thermal gradients, the segmentation network is treated strictly as the spatial front end, allowing the present work to focus entirely on the downstream depth measurement problem.

\begin{figure*}[!tb]
\centering
\includegraphics[width=6.5in]{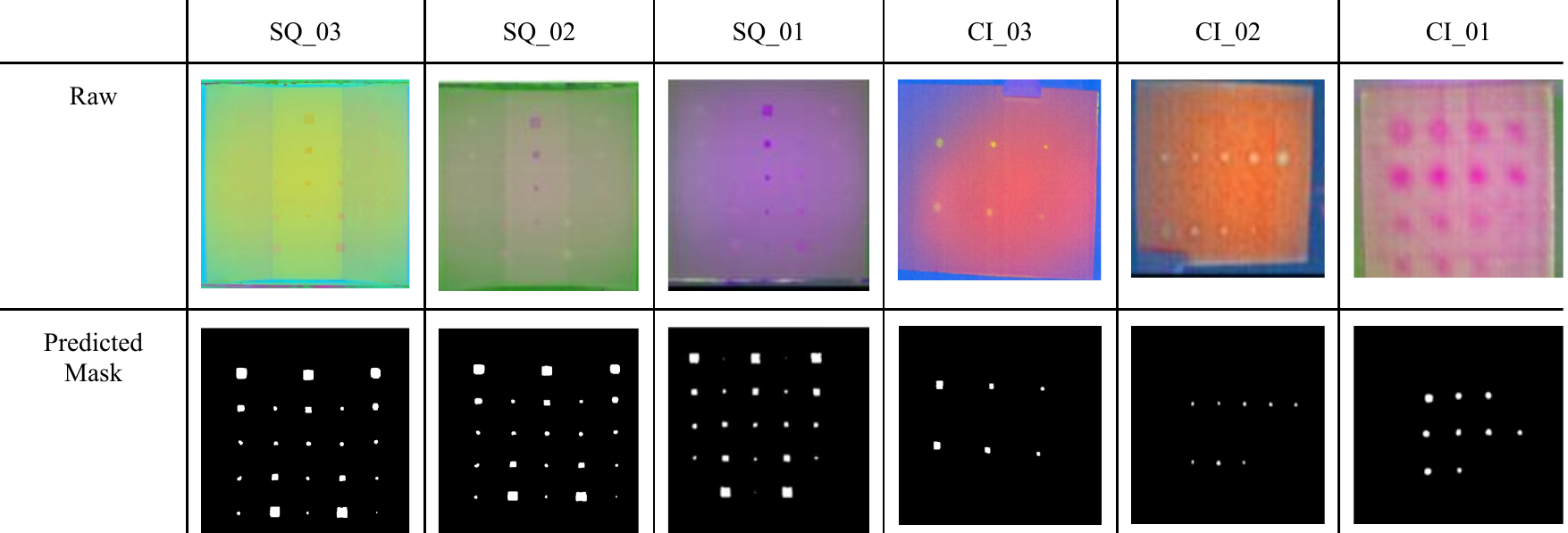}
\caption{Predicted binary defect masks for the six CFRP specimens obtained from the best segmentation architecture. Square and circular defects are reliably detected, providing spatial support for subsequent temporal measurement.}
\label{fig_masks}
\end{figure*}

The raw thermal sequence is processed by the trained segmentation network to produce a binary mask $\hat M$. Connected-component labeling then identifies individual defect regions
\begin{equation}
\label{eq:connected}
\{\Omega_k\}_{k=1}^{K}=\mathrm{ConnectedComponents}(\hat M),
\end{equation}
where $K$ is the number of detected defects. This representation allows square and circular specimens to be handled through the same downstream measurement procedure. Importantly, the $(x,y)$ position of a component is not supplied as a regression feature. The predicted masks for all six specimens are shown in Fig.~\ref{fig_masks}, demonstrating reliable detection across both geometries.

\subsection{Stage 2: Spatial-to-Temporal Decoupling}

Raw thermographic frames contain pixel-level camera noise and local thermal fluctuations. Passing all pixels directly to a regression network would preserve both useful thermal information and the geometric arrangement that can create spatial bias. Instead, the thermal response is averaged within each detected region:
\begin{equation}
\label{eq:spatial_avg}
T_k(t)=\frac{1}{|\Omega_k|}\sum_{(i,j)\in\Omega_k}X_{\mathrm{norm}}(i,j,t),
\end{equation}
where $T_k(t)$ is the normalized spatially averaged response, $|\Omega_k|$ is the number of pixels in defect region $\Omega_k$, and $X_{\mathrm{norm}}(i,j,t)$ is the globally normalized thermal intensity. The operation reduces local fluctuations and transforms each defect from a two-dimensional spatial signal into a one-dimensional temporal measurement. Under approximately independent pixel noise, averaging also reduces the standard deviation of the noise component approximately with $1/\sqrt{|\Omega_k|}$.

This operation is the central information-separation step. The regression model no longer sees whether a defect was located in the first or fifth column of a calibration grid. It sees the temporal response produced by the localized region. Spatial coordinates are retained only after regression for visualization and structural reconstruction.

\begin{figure*}[!tb]
\centering
\subfloat[SQ\_01 (0.2--1.0 mm)]{\includegraphics[width=3.2in]{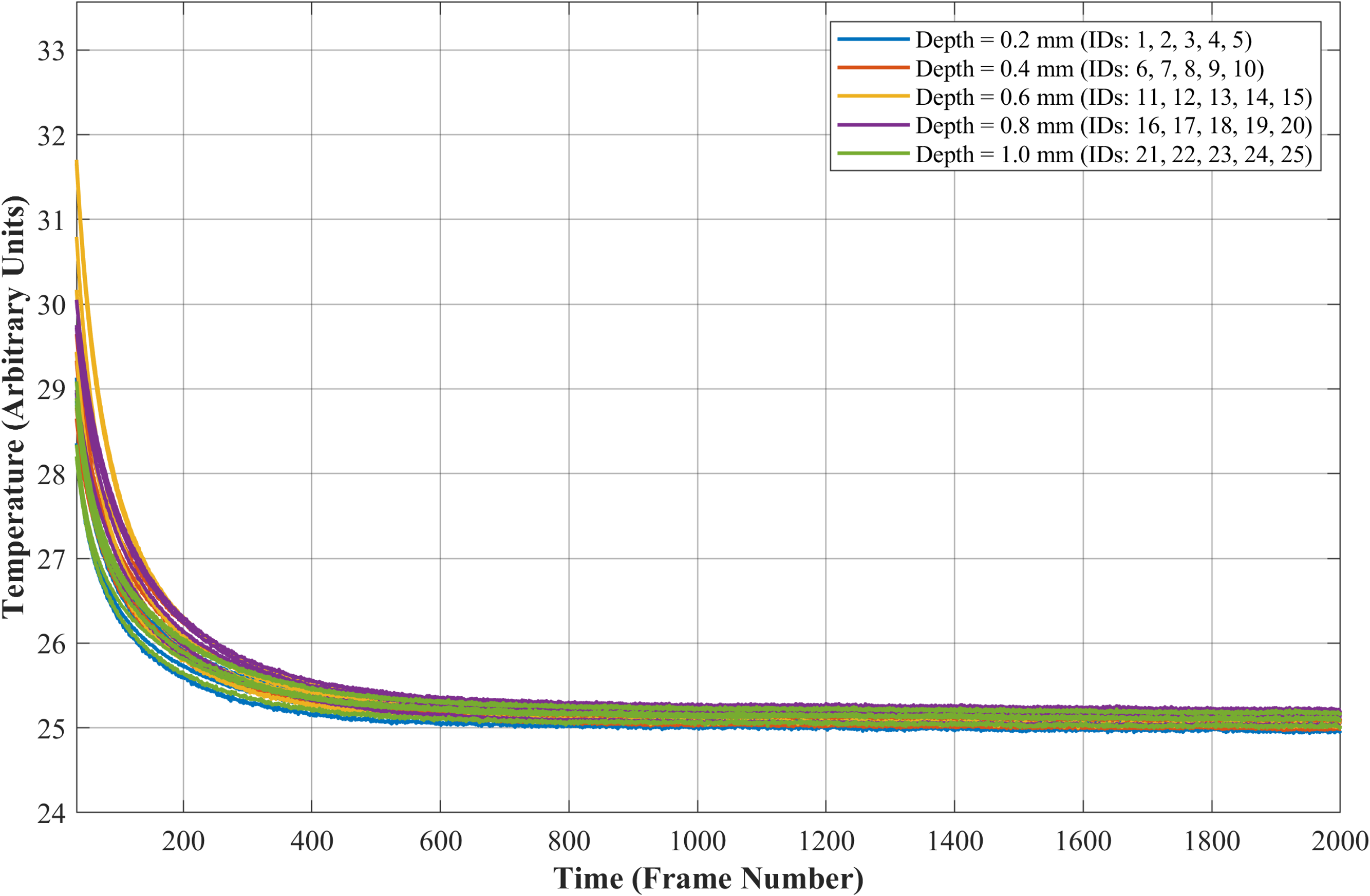}\label{fig_thermal_sq}}
\hfill
\subfloat[CI\_01 (2.0--2.2 mm)]{\includegraphics[width=3.2in]{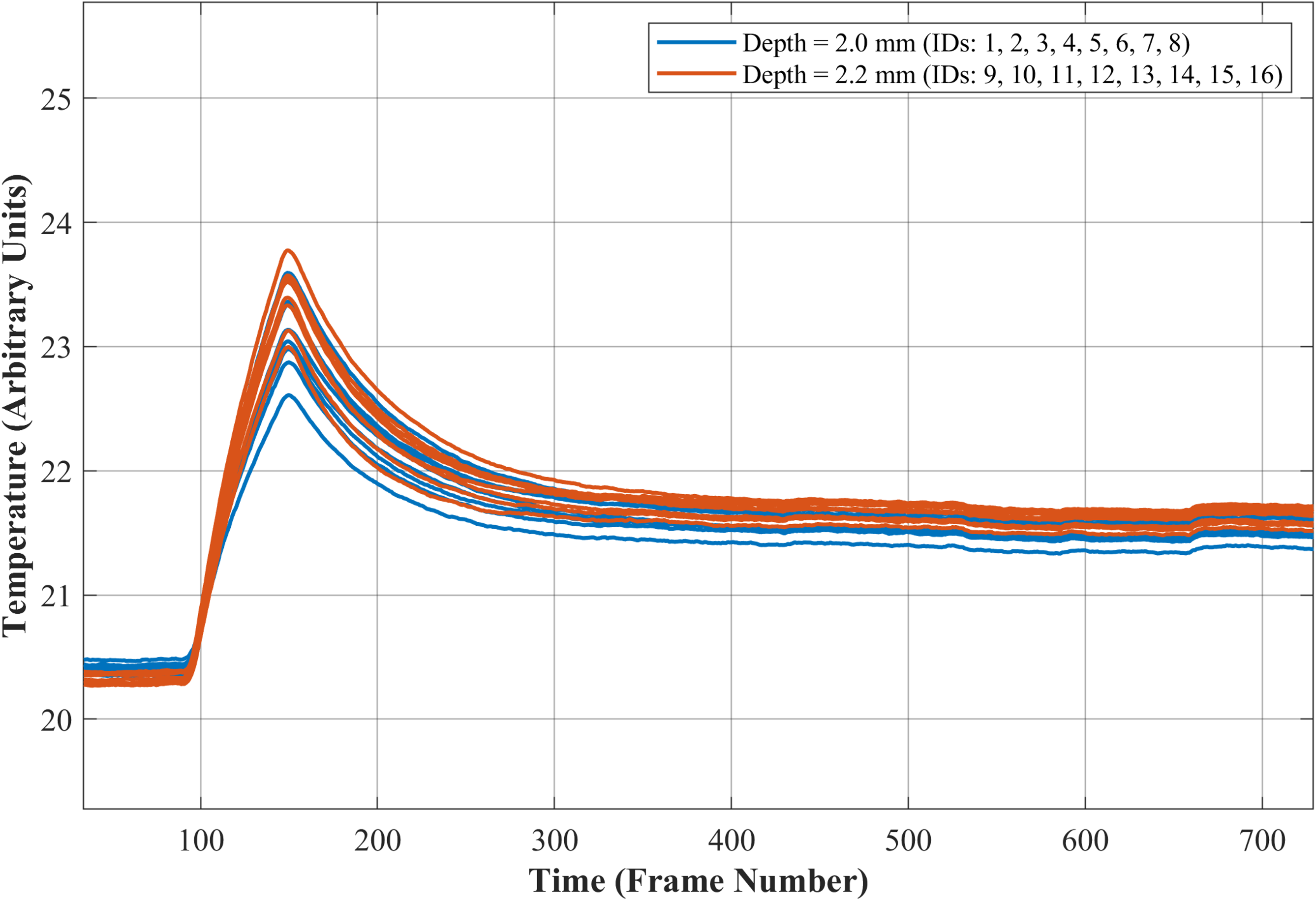}\label{fig_thermal_ci}}
\caption{Mean thermal responses after spatial averaging. Deeper defects exhibit delayed peak times and lower amplitudes, confirming that temporal features encode depth information.}
\label{fig_thermal_curves}
\end{figure*}

The measured curves in Fig.~\ref{fig_thermal_curves} illustrate the physical information retained by this compression. SQ\_01 contains five depth levels from 0.2 to 1.0 mm, while CI\_01 contains circular inserts at 2.0 and 2.2 mm. The depth levels remain distinguishable through changes in peak timing and thermal amplitude, demonstrating that spatial averaging does not remove the dominant depth information.

\subsection{Physics-Constrained Feature Representation}

A central requirement of the proposed representation is that the sixteen features should not be viewed as arbitrary dimensionality reduction. They impose a restricted measurement space in which the dominant depth information must come from the temporal heat response. The starting quantities are the inter-frame change
\begin{equation}
\label{eq:delta_t}
\Delta T_k(t)=T_k(t+1)-T_k(t),
\end{equation}
the temporal mean
\begin{equation}
\label{eq:mu_t}
\mu_T=\frac{1}{T}\sum_{t=1}^{T}T_k(t),
\end{equation}
and the centered response
\begin{equation}
\label{eq:tilde_t}
\widetilde T_k(t)=T_k(t)-\mu_T.
\end{equation}

The peak-time feature $t_{\mathrm{pk}}$ is directly linked to depth through \eqref{eq:heat_conduction}, while $d_{\mathrm{rel}}=\sqrt{t_{\mathrm{pk}}}$ follows from the same relationship after removing the constant factor $\beta\sqrt{\alpha}$. The remaining temporal and energy features characterize the amount and rate of heat retained during the acquisition window. Statistical features describe the shape and smoothness of the decay, while the limited geometric descriptors characterize defect extent rather than absolute image position. These sixteen variables, detailed in Table~\ref{tab:features}, act as strict mathematical constraints that expose the one-dimensional heat-conduction relationship. Thus, the feature vector contains information that can help correct for variations in the thermal response without explicitly exposing the calibration-grid coordinates.

\begin{table*}[!tb]
\caption{Sixteen Physics-Informed Features Used for Depth Regression}
\label{tab:features}
\centering
\begin{tabular}{@{}llc p{8.0cm}@{}}
\toprule
\textbf{Category} & \textbf{Feature} & \textbf{Formula} & \textbf{Physical / measurement interpretation}\\
\midrule
\multirow{5}{*}{Temporal}
& $t_{\mathrm{pk}}$ & $\arg\max_t T_k(t)$ & Peak time; primary depth-related temporal measurement through \eqref{eq:heat_conduction}.\\
& $\tau_r$ & $t_{\mathrm{pk}}-t_{\mathrm{onset}}$ & Rise duration associated with the transient response.\\
& $T_{\max}$ & $\max_t T_k(t)$ & Maximum surface temperature within the response window.\\
& $\dot T_{\max}$ & $\max_t \Delta T_k(t)/\Delta t$ & Maximum thermal change rate.\\
& $d_{\mathrm{rel}}$ & $\sqrt{t_{\mathrm{pk}}}$ & Linearized depth proxy derived from \eqref{eq:heat_conduction}.\\
\midrule
\multirow{3}{*}{Energy}
& $A_{\mathrm{auc}}$ & $\sum_{t=1}^{T}T_k(t)\Delta t$ & Integrated thermal response over the acquisition window.\\
& $E$ & $\sum_{t=1}^{T}T_k(t)^2\Delta t$ & Signal energy associated with the thermal decay.\\
& $T_{\mathrm{tail}}$ & $\mathrm{mean}(T_k(t)),\ t\in[0.8T,T]$ & Late-time thermal response, sensitive to persistent thermal contrast.\\
\midrule
\multirow{4}{*}{Statistical}
& $\sigma_T$ & $\sqrt{\frac{1}{T}\sum_t\widetilde T_k(t)^2}$ & Spread of the thermal response.\\
& $\gamma_1$ & $\frac{1}{T\sigma_T^3}\sum_t\widetilde T_k(t)^3$ & Skewness of the temporal response.\\
& $\gamma_2$ & $\frac{1}{T\sigma_T^4}\sum_t\widetilde T_k(t)^4-3$ & Excess kurtosis and tail weight.\\
& $\rho$ & $\frac{\sum_t\widetilde T_k(t)\widetilde T_k(t+1)}{\sum_t\widetilde T_k(t)^2}$ & Lag-1 temporal autocorrelation and response smoothness.\\
\midrule
\multirow{4}{*}{Geometric}
& $A_k$ & $|\Omega_k|$ & Defect area in pixels.\\
& $D_k$ & $2\sqrt{A_k/\pi}$ & Equivalent circular diameter.\\
& $AR_k$ & $W_{\mathrm{bb}}/H_{\mathrm{bb}}$ & Bounding-box aspect ratio.\\
& $d_{\mathrm{rel,g}}$ & $\sqrt{t_{\mathrm{pk}}}$ & Peak-time depth-linking representation used across specimen groups.\\
\bottomrule
\end{tabular}
\end{table*}

These sixteen variables act as strict mathematical constraints on the regression problem. The model is not asked to infer depth from an unrestricted spatial tensor. Instead, it must explain the target through a compact representation containing quantities explicitly connected to heat diffusion and thermal decay. This is especially important for the Adv-MLP control: a high-capacity network can fit a small calibration set, but its capacity does not guarantee that it will discover the correct invariant. The constrained feature space reduces the degrees of freedom available for geometric memorization before the regression algorithm is applied.

\subsection{Algorithmic Diagnostics and Regularized Regression}

Four regression algorithms are deliberately retained throughout the study: Random Forest (RF), Gradient Boosting Machine (GBM), Advanced Multi-Layer Perceptron (Adv-MLP), and XGBoost. The comparison is diagnostic rather than simply a search for the lowest training error. RF and GBM serve as unregularized tree-based controls, Adv-MLP tests whether neural-network representation capacity can overcome the distribution shift, and XGBoost tests whether explicit regularization improves physical generalization.

RF and GBM are used as unregularized tree-based controls \cite{ref28}. Their matched-geometry performance is therefore treated as a calibration control rather than evidence of geometry-invariant physical learning.

The Adv-MLP is a feedforward network with three hidden layers of 128, 64, and 32 neurons, together with batch normalization and dropout. It is optimized with Adam using Huber loss and early stopping based on validation error. The Huber objective reduces sensitivity to large residuals, but it does not prevent a high-capacity network from fitting a small spatially structured dataset. The Adv-MLP is therefore an important control for separating robustness of the loss function from robustness of the representation.

XGBoost is the regularized model used for the final depth measurement \cite{ref29}. At boosting iteration $m$, its second-order objective is written as
\begin{equation}
\label{eq:xgboost_obj}
\mathcal{L}^{(m)}=\sum_{n=1}^{N}\left[g_n f_m(\mathbf{x}_n)+\frac{1}{2}h_n f_m(\mathbf{x}_n)^2\right]+\Omega(f_m),
\end{equation}
where $f_m$ is the tree added at iteration $m$, $\mathbf{x}_n$ is the sixteen-dimensional feature vector of defect $n$, and $g_n$ and $h_n$ are the first- and second-order gradients of the loss with respect to the current prediction. The regularization term is
\begin{equation}
\label{eq:xgboost_reg}
\Omega(f_m)=\alpha\lVert\mathbf{w}\rVert_1+\frac{1}{2}\lambda\lVert\mathbf{w}\rVert_2^2,
\end{equation}
where $\mathbf{w}$ contains the tree leaf weights and $\alpha$ and $\lambda$ control the L1 and L2 penalties. Column sampling is used together with these penalties. The intended effect is to discourage dependence on a small set of idiosyncratic feature dimensions and to reduce the capacity of the boosted ensemble to reproduce accidental training-set structure.

Thus, the two constraints operate at different levels: spatial decoupling removes the main geometric shortcut before regression, while L1/L2 regularization and column sampling control the remaining model capacity.

\subsection{Feature Diagnostic}

\begin{figure}[!tb]
\centering
\includegraphics[width=3.5in]{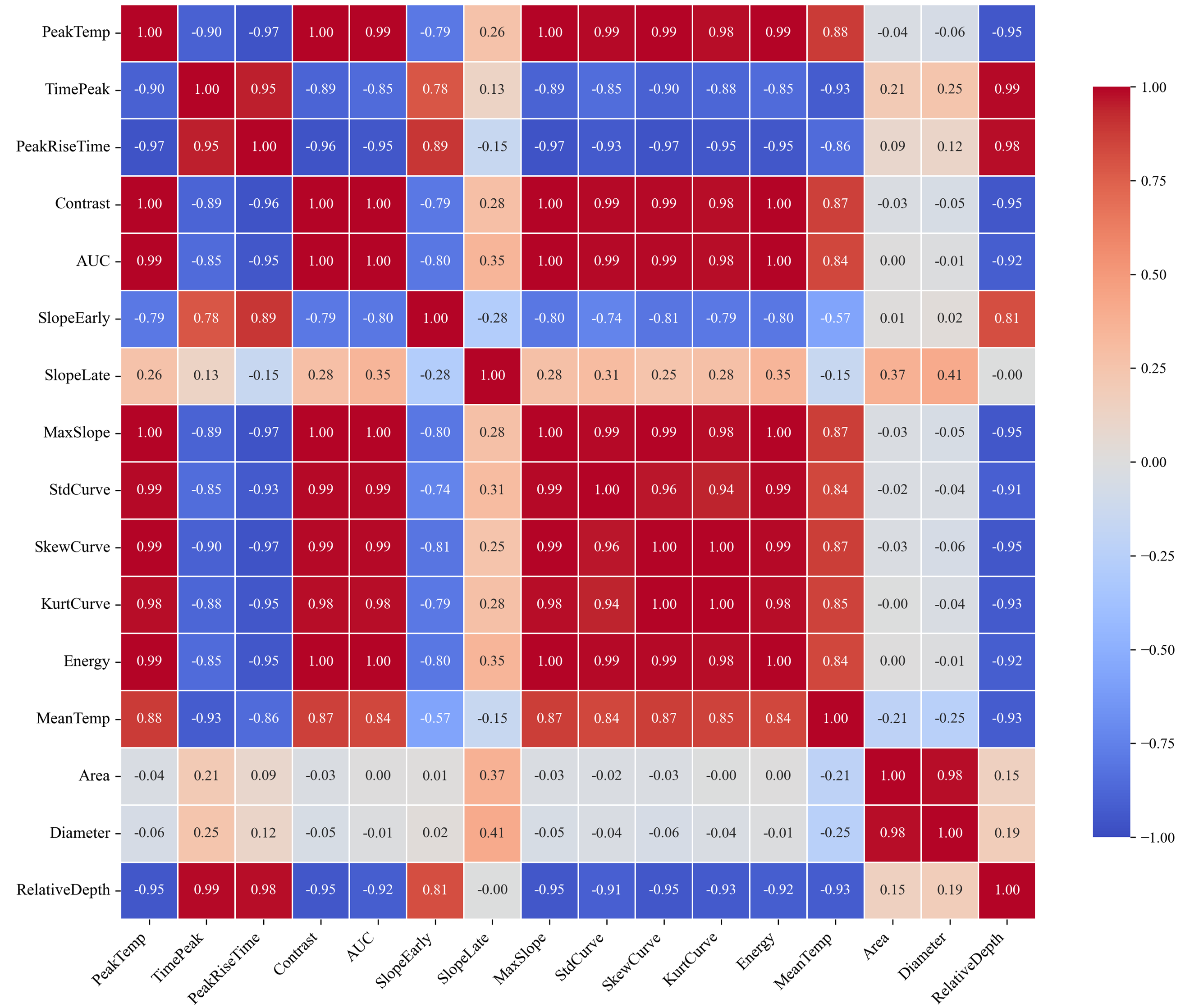}
\caption{Pearson correlation matrix for the sixteen features and depth. Temporal features exhibit the strongest correlation, confirming the physics-informed design.}
\label{fig_correlation_matrix}
\end{figure}

The Pearson correlation matrix, presented in Fig.~\ref{fig_correlation_matrix}, provides an independent diagnostic of the feature space. The temporal variables, particularly $t_{\mathrm{pk}}$ and $d_{\mathrm{rel}}$, show the strongest relationship with depth, whereas geometric variables are weaker indicators of through-thickness position. This is consistent with \eqref{eq:heat_conduction}: depth is primarily encoded in the temporal diffusion response, while defect area and aspect ratio describe lateral extent and shape.

\subsection{Automated 3D Reconstruction for Structural Analysis}
\label{sec:3d_recon}

The final step converts the predicted depth $\hat{d}_k$ into a three-dimensional surface mesh. Pixel coordinates are converted to physical units via the spatial calibration factor:
\begin{equation}
\label{eq:calibration}
x_{\mathrm{mm}}=j\,s_{\mathrm{px}},\qquad y_{\mathrm{mm}}=i\,s_{\mathrm{px}},\qquad s_{\mathrm{px}}=\frac{L_{\mathrm{spec}}}{N_{\mathrm{px}}}.
\end{equation}
For Dataset 1, the calibration factor $s_{\mathrm{px}}$ is 0.586 mm/pixel; for Dataset 2, it is 0.391 mm/pixel. Each defect region $\Omega_k$ is treated as a thin delamination with top and bottom surfaces:
\begin{align}
\label{eq:surface_top}
Z_{\mathrm{top}}(x,y) &= \hat{d}_k, \quad &&\forall(x,y)\in\Omega_k,\\
\label{eq:surface_bot}
Z_{\mathrm{bot}}(x,y) &= \hat{d}_k+\delta_{\mathrm{gap}}, \quad &&\forall(x,y)\in\Omega_k,
\end{align}
where $\delta_{\mathrm{gap}}$ represents the 0.0075 mm Teflon insert thickness. Point clouds for the top and bottom surfaces are generated as:
\begin{align}
\label{eq:cloud_top}
\mathbf{P}^{\mathrm{top}}_k &= (j s_{\mathrm{px}}, i s_{\mathrm{px}}, \hat{d}_k),\\
\label{eq:cloud_bot}
\mathbf{P}^{\mathrm{bot}}_k &= (j s_{\mathrm{px}}, i s_{\mathrm{px}}, \hat{d}_k+\delta_{\mathrm{gap}}).
\end{align}
Delaunay triangulation \cite{ref30} is applied to the $xy$-coordinates of each point cloud to produce the surface meshes $\mathcal{M}_k$. The mean surface distance for defect $k$ equals the absolute regression error, given by $\mathrm{MSD}_k=|\hat{d}_k-d_k^{\mathrm{true}}|$. The 3D centroid is obtained from the segmentation centroid and predicted depth, enabling automated spatial localization. Because this step operates on localized depth estimates rather than full-volume tensors, reconstruction completes in 3--5 seconds per specimen on standard hardware.

\medskip
\noindent Algorithm \ref{alg:framework} summarizes the complete spatio-temporal decoupled measurement framework, integrating segmentation, feature extraction, regression, and 3D reconstruction.

\begin{algorithm}[!tb]
\caption{Spatio-Temporal Decoupled Computational Measurement Framework}
\label{alg:framework}
\begin{algorithmic}[1]
\REQUIRE Raw thermal tensor $\mathbf{X}\in\mathbb{R}^{H\times W\times T}$; trained segmentation network; training defects with depths $\{d_k\}$
\ENSURE Cross-validated predictions for RF, GBM, Adv-MLP, XGBoost; final depths $\hat{d}_k$; 3D meshes $\mathcal{M}_k$
\STATE $\hat M\leftarrow\mathrm{SegmentationNet}(\mathbf{X})$
\STATE $\{\Omega_k\}_{k=1}^{K}\leftarrow\mathrm{ConnectedComponents}(\hat M)$
\FOR{$k=1$ \TO $K$}
    \STATE $T_k(t)\leftarrow\frac{1}{|\Omega_k|}\sum_{(i,j)\in\Omega_k}X_{\mathrm{norm}}(i,j,t)$
    \STATE Compute $\Delta T_k(t)$, $\mu_T$, and $\widetilde T_k(t)$
    \STATE $\mathbf{f}_k\leftarrow$ sixteen temporal, energy, statistical, and geometric features
    \STATE Standardize $\mathbf{f}_k$ using training-set statistics only
\ENDFOR
\STATE Partition defects by specimen according to the predefined fold
\STATE Train/evaluate RF, GBM, and Adv-MLP on the same $\mathbf{f}_k$
\STATE Train/evaluate regularized XGBoost using \eqref{eq:xgboost_obj}--\eqref{eq:xgboost_reg}
\FOR{each test defect $k$}
    \STATE $\hat{d}_k\leftarrow f_{\mathrm{XGBoost}}(\mathbf{f}_k)$
    \STATE $\mathbf{P}^{\mathrm{top}}_k, \mathbf{P}^{\mathrm{bot}}_k \leftarrow$ point clouds at $\hat{d}_k$ and $\hat{d}_k+\delta_{\mathrm{gap}}$
    \STATE $\mathcal{M}_k\leftarrow\mathrm{DelaunayTriangulation}(\mathbf{P}^{\mathrm{top}}_k,\mathbf{P}^{\mathrm{bot}}_k)$
\ENDFOR
\RETURN $\{\hat{d}_k,\mathcal{M}_k\}$ and fold-wise regression metrics
\end{algorithmic}
\end{algorithm}

% ==========================================
% SECTION III: VALIDATION
% ==========================================
\section{Experimental Validation}
\label{sec:validation}

\subsection{Datasets and Specimen-Level Cross-Validation}

Six CFRP specimens from two public databases are used \cite{ref31,ref32}. The two databases are valuable for this diagnostic experiment because they provide different specimen geometries, defect arrangements, spatial resolutions, and depth ranges. The purpose is not to merge them into one homogeneous dataset, but to determine whether a model trained on one spatial organization preserves its calibration when the organization changes.

Dataset 1 is derived from the thermal imaging dataset of Erazo-Aux et al. \cite{ref31}. It contains three specimens denoted here as SQ\_01, SQ\_02, and SQ\_03, corresponding to CFRP006, CFRP007, and CFRP008 in the original dataset. They represent flat, curved, and trapezoidal specimen geometries, respectively. Each specimen contains 25 square Teflon inserts arranged in a $5\times5$ grid. The insert depths are 0.2, 0.4, 0.6, 0.8, and 1.0 mm. The spatial calibration factor $s_{\mathrm{px}}$ is 0.586 mm/pixel for a 512-pixel image over a 300 mm specimen edge.

Dataset 2 is derived from the CFRP thermal-imaging work of Ahmed et al. \cite{ref32}. It contains CI\_01, CI\_02, and CI\_03, corresponding to specimens 3, 1, and 2 in the original dataset. These specimens contain circular inserts with diameters of 2, 4, 6, and 8 mm at two depth levels, 2.0 and 2.2 mm. The spatial calibration factor is 0.391 mm/pixel. The circular specimens are particularly useful for diagnosing spatial bias because they do not reproduce the rigid five-column arrangement of the square-insert specimens.

The specimen-level cross-validation partition is shown in Table~\ref{tab:partition}. The numbers in parentheses denote the available defects used in each partition after localization. No individual defect from a specimen is split across training and test sets. This prevents a model from learning specimen-specific spatial or thermal signatures and then being evaluated on another region of the same specimen.

\begin{table}[!tb]
\caption{Three-Fold Specimen-Level Cross-Validation Partitioning}
\label{tab:partition}
\centering
\begin{tabular}{@{}c c c c@{}}
\toprule
\textbf{Fold} & \textbf{Training} & \textbf{Validation} & \textbf{Test}\\
\midrule
1 & SQ\_01, CI\_01 (41) & SQ\_02, CI\_02 (35) & SQ\_03, CI\_03 (31)\\
2 & SQ\_03, CI\_03 (31) & SQ\_01, CI\_01 (41) & SQ\_02, CI\_02 (35)\\
3 & SQ\_02, CI\_02 (35) & SQ\_03, CI\_03 (31) & SQ\_01, CI\_01 (41)\\
\bottomrule
\end{tabular}
\end{table}

Fold 2 is the most important diagnostic split. Its training data include SQ\_03, a square-insert specimen with a rigid five-column arrangement, while the test set includes CI\_02, where the circular defect arrangement does not provide the same column structure. A model that uses geometry as a hidden proxy for depth should therefore suffer a large calibration change in this fold.

\subsection{Evaluation Metrics}

Three metrics are reported. Mean absolute error is the primary measurement metric because it is expressed directly in millimeters:
\begin{equation}
\label{eq:mae}
\mathrm{MAE} = \frac{1}{N}\sum_{n=1}^{N}|d_n - \hat{d}_n|,
\end{equation}
where $d_n$ and $\hat{d}_n$ are the true and predicted depths. Root mean square error is
\begin{equation}
\label{eq:rmse}
\mathrm{RMSE} = \sqrt{\frac{1}{N}\sum_{n=1}^{N}(d_n - \hat{d}_n)^2},
\end{equation}
which assigns greater influence to large calibration errors. The coefficient of determination is
\begin{equation}
\label{eq:r2}
R^2 = 1 - \frac{\sum_{n=1}^{N}(d_n - \hat{d}_n)^2}{\sum_{n=1}^{N}(d_n - \bar{d})^2}, \quad \text{where } \bar{d} = \frac{1}{N}\sum_{n=1}^{N}d_n.
\end{equation}
For the circular specimens, the $R^2$ performance must be interpreted together with MAE and RMSE because the target range is narrow. A modest absolute error can produce a strongly negative $R^2$ when the variance of the true labels is small.

% ==========================================
% SECTION IV: RESULTS
% ==========================================
\section{Results and Discussion}
\label{sec:results}

The results are interpreted primarily as a generalization diagnostic. High accuracy on a matched geometry is not sufficient evidence that a model has learned thermal physics. The decisive evidence is whether the same calibration survives a change in specimen geometry and defect arrangement.

\begin{figure*}[!tb]
\centering
\subfloat[Fold 1 ($R^2=0.992$)]{\includegraphics[width=0.28\linewidth]{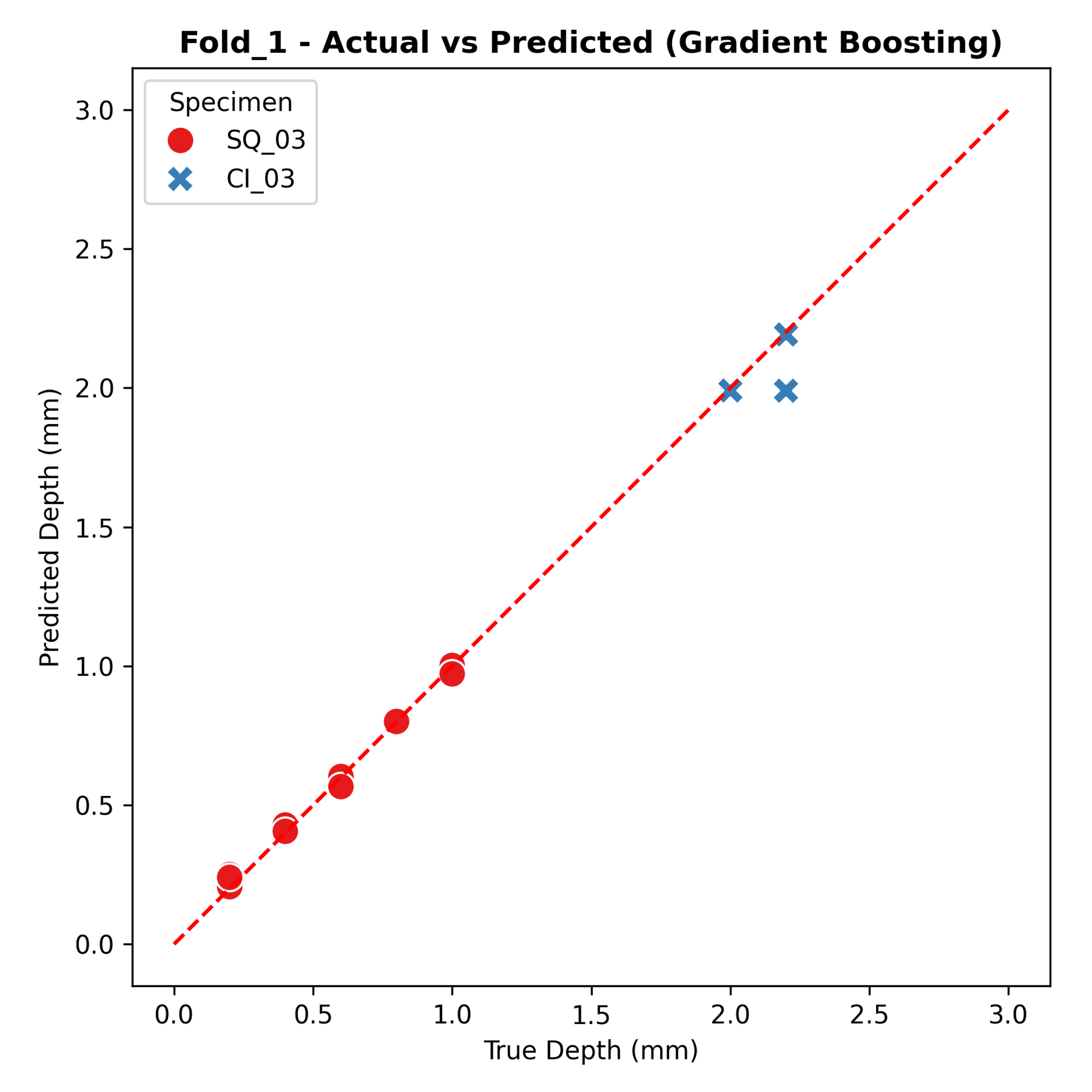}}
\hfill
\subfloat[Fold 2 ($R^2=0.992$)]{\includegraphics[width=0.28\linewidth]{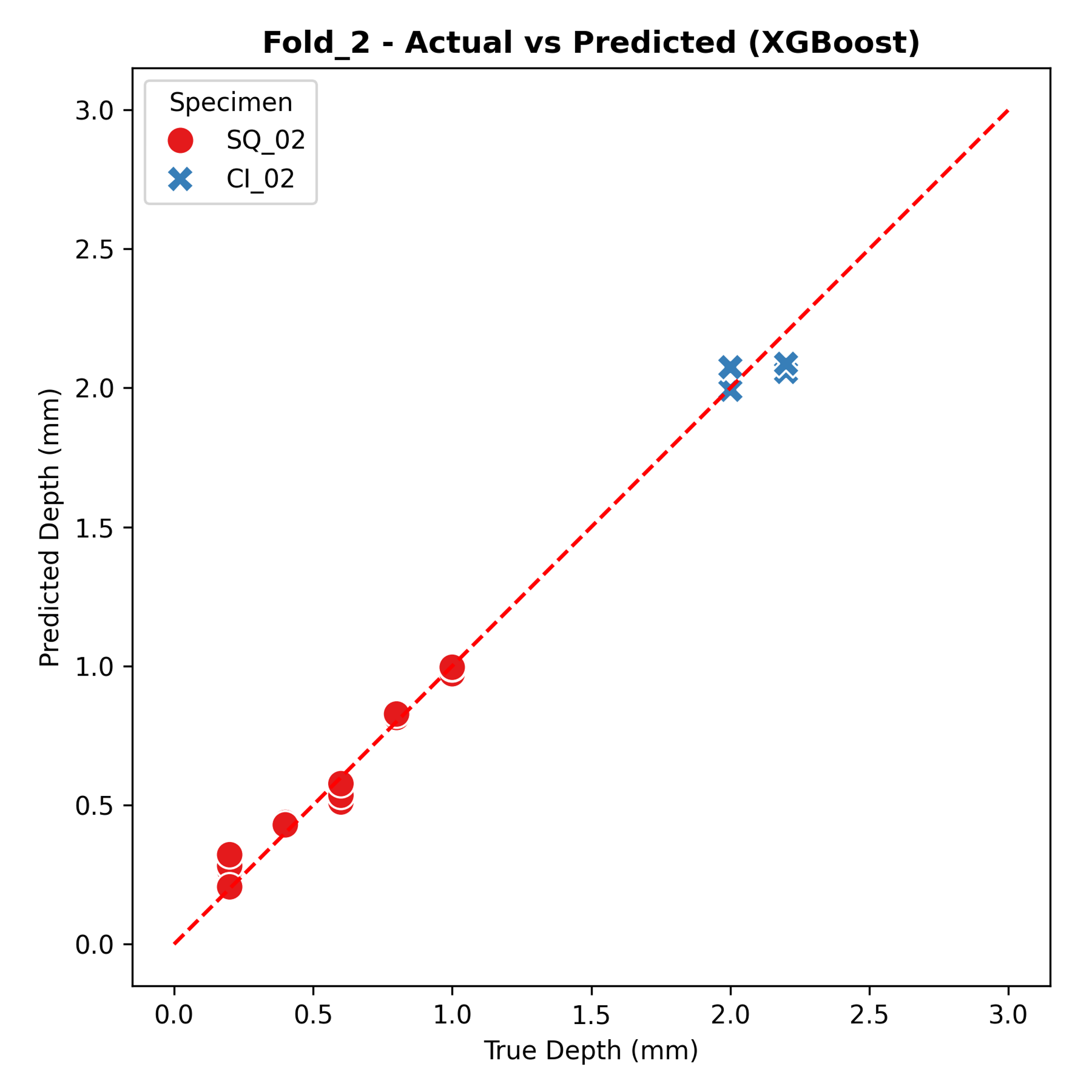}}
\hfill
\subfloat[Fold 3 ($R^2=0.971$)]{\includegraphics[width=0.28\linewidth]{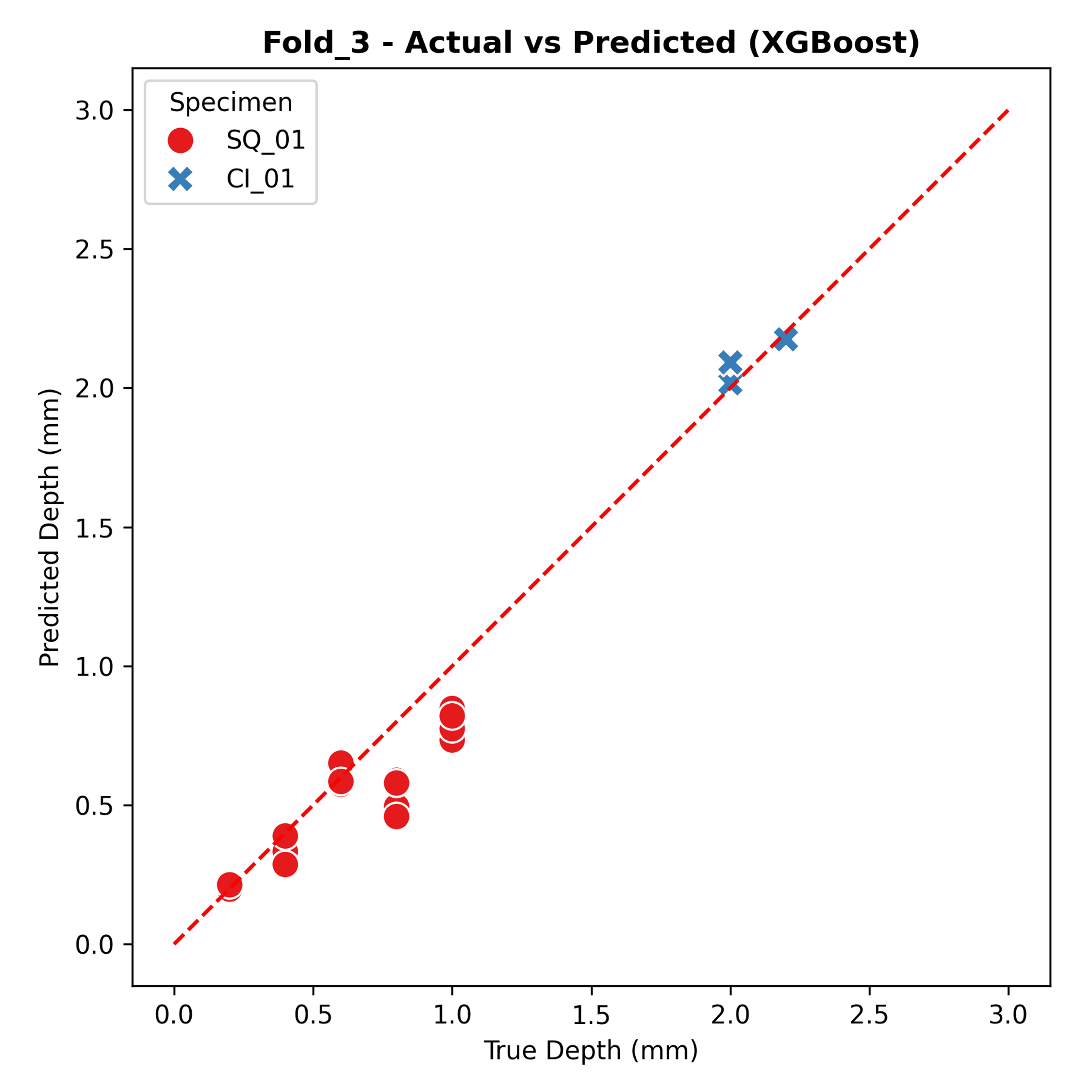}}
\caption{Scatter plots of actual versus predicted depth across the three specimen-level cross-validation folds. Fold 1 (a) displays the calibration of the Gradient Boosting control, while Fold 2 (b) and Fold 3 (c) demonstrate the robust generalization of the regularized XGBoost model under distribution shifts from rigid square grids to columnless circular layouts.}
\label{fig_scatter_folds}
\end{figure*}

Fig.~\ref{fig_scatter_folds} provides the visual fold-wise scatter plots comparing true versus predicted depths across the cross-validation splits. The numerical results across all folds are summarized in Table~\ref{tab:results}, with the best-performing metrics highlighted in bold.

\begin{table*}[!tb]
\caption{Depth Regression Results Across All Folds}
\label{tab:results}
\centering
\begin{tabular}{@{}ccccc ccc ccc ccc@{}}
\toprule
\textbf{Fold} & \textbf{Test} & \multicolumn{3}{c}{\textbf{RF}} & \multicolumn{3}{c}{\textbf{GBM}} & \multicolumn{3}{c}{\textbf{XGBoost}} & \multicolumn{3}{c}{\textbf{Adv-MLP}}\\
\cmidrule(lr){3-5}\cmidrule(lr){6-8}\cmidrule(lr){9-11}\cmidrule(lr){12-14}
& & MAE & RMSE & $R^2$ & MAE & RMSE & $R^2$ & MAE & RMSE & $R^2$ & MAE & RMSE & $R^2$\\
\midrule
\multirow{3}{*}{1}
& SQ\_03 & 0.0102 & 0.0171 & \textbf{0.996} & 0.0125 & 0.0184 & \textbf{0.996} & \textbf{0.0099} & \textbf{0.0169} & \textbf{0.996} & 0.1458 & 0.1725 & 0.628\\
& CI\_03 & 0.0647 & 0.1021 & $-0.042$ & \textbf{0.0400} & \textbf{0.0861} & \textbf{0.259} & 0.1154 & 0.1309 & $-0.714$ & 0.3385 & 0.3665 & $-12.43$\\
& Overall & 0.0207 & 0.0475 & 0.995 & \textbf{0.0178} & \textbf{0.0413} & \textbf{0.996} & 0.0303 & 0.0596 & 0.992 & 0.1831 & 0.2236 & 0.880\\
\midrule
\multirow{3}{*}{2}
& SQ\_02 & \textbf{0.0120} & \textbf{0.0158} & \textbf{0.997} & 0.0162 & 0.0199 & 0.995 & 0.0359 & 0.0461 & 0.973 & 0.5221 & 0.5730 & $-3.103$\\
& CI\_02 & 0.5209 & 0.5326 & $-27.37$ & 0.6834 & 0.7232 & $-51.29$ & \textbf{0.0756} & \textbf{0.0967} & \textbf{0.066} & 0.1370 & 0.1622 & $-1.632$\\
& Overall & 0.1574 & 0.2850 & 0.844 & 0.2068 & 0.3869 & 0.712 & \textbf{0.0473} & \textbf{0.0647} & \textbf{0.992} & 0.4121 & 0.4919 & 0.534\\
\midrule
\multirow{3}{*}{3}
& SQ\_01 & 0.1751 & 0.2437 & 0.257 & 0.2215 & 0.2812 & 0.011 & \textbf{0.1192} & \textbf{0.1608} & \textbf{0.677} & 0.3899 & 0.4293 & $-1.303$\\
& CI\_01 & 0.0362 & 0.0414 & 0.829 & \textbf{0.0260} & \textbf{0.0292} & \textbf{0.915} & 0.0414 & 0.0527 & 0.722 & 0.8308 & 0.8356 & $-68.82$\\
& Overall & 0.1209 & 0.1921 & 0.937 & 0.1452 & 0.2204 & 0.917 & \textbf{0.0888} & \textbf{0.1298} & \textbf{0.971} & 0.5620 & 0.6204 & 0.346\\
\bottomrule
\end{tabular}
\end{table*}

\subsection{Cross-Fold Regression Performance}

Fold 1 represents the relatively favorable case in which training and test specimens preserve comparable square and circular families. RF and GBM therefore obtain low errors on several test groups. Their good performance in this fold is useful as a control, but it cannot establish that the models have learned a geometry-invariant physical relationship. In fact, the Adv-MLP already shows instability: its overall Fold 1 MAE is 0.1831 mm despite its greater representational capacity. This is consistent with the risk of fitting a small collection of discrete calibration defects rather than learning a stable thermal-depth mapping.

Fold 2 provides the strongest evidence of spatial dataset bias. Training includes SQ\_03, whose square inserts occupy a rigid five-column grid. The test set contains SQ\_02 and, critically, CI\_02. On CI\_02, RF reaches an MAE of 0.5209 mm and GBM reaches 0.6834 mm. These errors are larger than the 0.2 mm separation between the two circular depth levels. Their very negative $R^2$ values, $-27.37$ and $-51.29$, respectively, show that the models fail to preserve the calibration when the spatial proxy changes.

The key point is not that tree models are inherently unsuitable for thermographic measurement. Rather, the controlled experiment shows that unregularized tree ensembles can exploit correlations present in the training geometry. When the five-column arrangement disappears, the correlation between position and depth disappears as well, and the learned mapping becomes unstable. This is the characteristic signature of spatial overfitting.

XGBoost behaves differently in the same Fold 2 test. Its CI\_02 MAE is 0.0756 mm and its overall Fold 2 MAE is 0.0473 mm, achieving an $R^2$ of 0.992 for the overall test set. This does not mean that regularization magically creates physical information. Rather, the model is operating on a representation that has already removed pixel-level spatial coordinates, while the L1/L2 objective and column sampling constrain the remaining model capacity. The strongest depth-related variable, $d_{\mathrm{rel}}=\sqrt{t_{\mathrm{pk}}}$, therefore becomes a natural route for the model to preserve calibration across the geometric shift.

Fold 3 introduces a different distribution shift by training on the curved SQ\_02 specimen and testing on the flat SQ\_01 specimen. All models experience some degradation, showing that geometry changes can alter the measured thermal response even when the defect depths are the same. XGBoost retains an overall MAE of 0.0888 mm and an $R^2$ of 0.971, whereas the Adv-MLP reaches 0.5620 mm overall and exhibits a strongly negative $R^2$ on CI\_01. The result indicates that increasing neural-network capacity does not automatically improve physical generalization.

Averaged across the three folds, regularized XGBoost achieves an MAE of 0.056 mm and an RMSE of 0.085 mm. The corresponding overall GBM error is approximately 0.123 mm. Thus, the advantage of XGBoost is not based on a single favorable split. More importantly, the regularized model maintains substantially better absolute calibration in the geometric distribution-shift case that exposes the weakness of the unregularized controls.

\begin{figure*}[!tb]
\centering
\subfloat[SQ\_03: 25 square defects, 0.2--1.0 mm depth]{\includegraphics[width=0.44\linewidth]{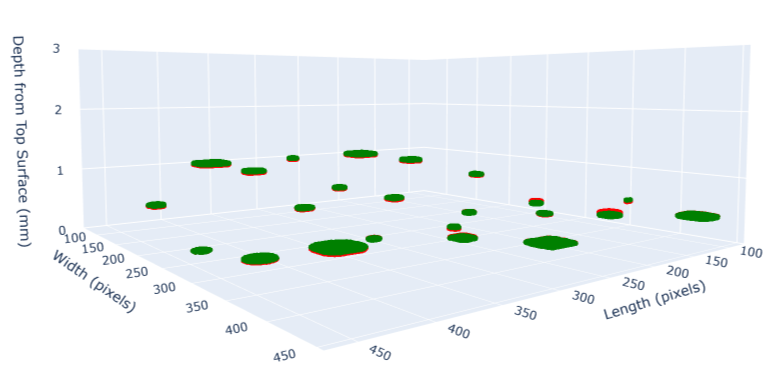}}
\hfill
\subfloat[CI\_03: 6 circular defects, 2.0 and 2.2 mm depth]{\includegraphics[width=0.44\linewidth]{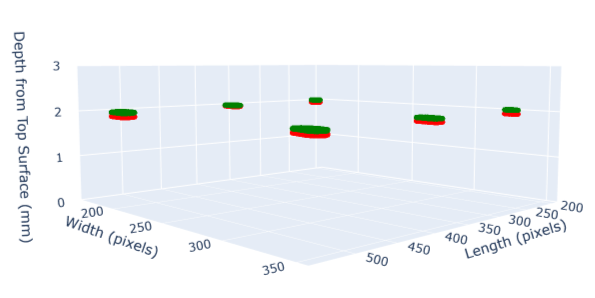}}
\caption{Three-dimensional structural reconstructions from XGBoost depth predictions. Green surfaces denote the true delamination boundaries, while red surfaces represent the computational predictions, demonstrating accurate spatial registration and clear resolution of closely spaced layers.}
\label{fig_3d_recon}
\end{figure*}

\subsection{Three-Dimensional Computational Reconstruction}

The three-dimensional output is shown in Fig.~\ref{fig_3d_recon}. The reconstruction is performed using the method described in Section \ref{sec:3d_recon}, applying Delaunay triangulation to the top and bottom point clouds generated from the predicted depths and segmentation masks. In the visual rendering, green surfaces represent the ground truth defect geometries, while red surfaces illustrate the computational depth estimates predicted by the regularized XGBoost model. The process completes in 3--5 seconds per specimen on standard workstation hardware, and the mean surface distance for each defect equals the regression MAE, as every pixel within a localized defect receives the same predicted depth.

The reconstruction of SQ\_03 (Fig.~\ref{fig_3d_recon}, left) resolves all 25 square defects across the five depth layers (0.2, 0.4, 0.6, 0.8, and 1.0 mm), with close spatial alignment between predicted (red) and true (green) defect positions. For CI\_03 (Fig.~\ref{fig_3d_recon}, right), the 0.2 mm spacing between the 2.0 and 2.2 mm circular defect levels is distinctly visible, confirming that the pipeline preserves sub-millimeter depth resolution. The visual overlay provides an intuitive view of the defect distribution, which is directly usable for finite element analysis or digital twin integration. Because the pipeline operates on localized physics features rather than full-volume tensors, it avoids the memory and time overhead of volumetric reconstruction methods.

% ==========================================
% SECTION V: CONCLUSION
% ==========================================
\section{Conclusion}
\label{sec:conclusion}

This work presented a spatio-temporal decoupling architecture for computational depth measurement from thermographic video of CFRP subsurface defects. The results show that regular-grid calibration can create spatial dataset bias: RF, GBM, and the over-parameterized Adv-MLP can fit matched layouts but degrade under geometric shifts. By restricting regression to spatially averaged temporal responses, the framework removes the main geometric shortcut before learning.

Sixteen physics-informed features impose explicit mathematical constraints related to heat diffusion, while regularized XGBoost adds L1/L2 penalties and column sampling. Under specimen-level cross-validation, the combined approach achieves an MAE of 0.056 mm and an RMSE of 0.085 mm. Fold 2 is the clearest evidence: RF and GBM collapse on the columnless circular layout, whereas XGBoost retains substantially lower absolute error.

The framework assumes one depth per localized defect and is therefore not yet suited to sloped or multi-level impact damage. Small defects near the camera resolution limit and segmentation errors remain practical constraints. Future work will investigate spatially varying depth estimation while preserving the same protection against spatial overfitting and will extend validation to real impact damage.

\balance
{\fontsize{8}{9}\selectfont

}


\begin{thebibliography}{32}
\setlength{\itemsep}{1pt}
\bibliographystyle{IEEEtran}

\bibitem{ref1}
L. Mishnaevsky, K. Branner, H. N. Petersen, J. Beauson, M. McGugan, and B. F. S{\o}rensen, ``Materials for wind turbine blades: An overview,'' \textit{Materials}, vol. 10, no. 11, p. 1285, 2017. doi: \href{https://doi.org/10.3390/ma10111285}{10.3390/ma10111285}.

\bibitem{ref2}
S. Abrate, \textit{Impact on Composite Structures}. Cambridge, U.K.: Cambridge University Press, 1998. doi: \href{https://doi.org/10.1017/CBO9780511574504}{10.1017/CBO9780511574504}.

\bibitem{ref3}
C. Kassapoglou, \textit{Design and Analysis of Composite Structures: With Applications to Aerospace Structures}, 2nd ed. Chichester, U.K.: Wiley, 2013. doi: \href{https://doi.org/10.1002/9781118536933}{10.1002/9781118536933}.

\bibitem{ref4}
X. P. Maldague, \textit{Theory and Practice of Infrared Technology for Nondestructive Testing}. New York, NY, USA: Wiley, 2001.

\bibitem{ref5}
C. Meola and G. M. Carlomagno, ``Recent advances in the use of infrared thermography,'' \textit{Meas. Sci. Technol.}, vol. 15, no. 9, pp. R27--R58, 2004. doi: \href{https://doi.org/10.1088/0957-0233/15/9/R01}{10.1088/0957-0233/15/9/R01}.

\bibitem{ref6}
H. S. Carslaw and J. C. Jaeger, \textit{Conduction of Heat in Solids}, 2nd ed. Oxford, U.K.: Oxford University Press, 1959.

\bibitem{ref7}
S. M. Shepard, J. R. Lhota, B. A. Rubadeux, D. Wang, and T. Ahmed, ``Reconstruction and enhancement of active thermographic image sequences,'' \textit{Opt. Eng.}, vol. 42, no. 5, pp. 1337--1342, 2003. doi: \href{https://doi.org/10.1117/1.1566969}{10.1117/1.1566969}.

\bibitem{ref8}
A. Schager, G. Zauner, G. Mayr, and P. Burgholzer, ``Extension of the thermographic signal reconstruction technique for an automated segmentation and depth estimation of subsurface defects,'' \textit{J. Imaging}, vol. 6, no. 9, p. 96, 2020. doi: \href{https://doi.org/10.3390/jimaging6090096}{10.3390/jimaging6090096}.

\bibitem{ref9}
C. Ibarra-Castanedo and X. Maldague, ``Pulsed phase thermography reviewed,'' \textit{Quant. InfraRed Thermogr. J.}, vol. 1, no. 1, pp. 47--70, 2004. doi: \href{https://doi.org/10.3166/qirt.1.47-70}{10.3166/qirt.1.47-70}.

\bibitem{ref10}
N. Rajic, ``Principal component thermography for flaw contrast enhancement and flaw depth characterisation in composite structures,'' \textit{Compos. Struct.}, vol. 58, no. 4, pp. 521--528, 2002. doi: \href{https://doi.org/10.1016/S0263-8223(02)00161-7}{10.1016/S0263-8223(02)00161-7}.

\bibitem{ref11}
S. Ekanayake, S. Gurram, and R. H. Schmitt, ``Depth determination of defects in CFRP-structures using lock-in thermography,'' \textit{Compos. Part B: Eng.}, vol. 147, pp. 128--134, 2018. doi: \href{https://doi.org/10.1016/j.compositesb.2018.04.032}{10.1016/j.compositesb.2018.04.032}.

\bibitem{ref12}
Z. Wei, H. Fernandes, H.-G. Herrmann, J. R. Tarpani, and A. Osman, ``A deep learning method for the impact damage segmentation of curve-shaped CFRP specimens inspected by infrared thermography,'' \textit{Sensors}, vol. 21, no. 2, p. 395, 2021. doi: \href{https://doi.org/10.3390/s21020395}{10.3390/s21020395}.

\bibitem{ref13}
H. Memon, Z. U. Abidin, J. Ahmed, and A. Siddiqua, ``Debond Detection in CFRP Structures using Optical Pulse Thermography and Deep Learning,'' in \textit{Proc. 5th Int. Conf. Comput. Math. Eng. Technol. (iCoMET)}, Sukkur, Pakistan, 2026, pp. 1--6. doi: \href{https://doi.org/10.1109/iCoMET69771.2026.11591843}{10.1109/iCoMET69771.2026.11591843}.

\bibitem{ref14}
L. Santoro and R. Sesana, ``A physics-informed framework for feature extraction and defect segmentation in pulsed infrared thermography,'' \textit{Eng. Failure Anal.}, vol. 175, art. 109542, 2025. doi: \href{https://doi.org/10.1016/j.engfailanal.2025.109542}{10.1016/j.engfailanal.2025.109542}.

\bibitem{ref15}
Y. Dong, C. Xia, J. Yang, Y. Cao, and X. Li, ``Spatio-temporal 3-D residual networks for simultaneous detection and depth estimation of CFRP subsurface defects in lock-in thermography,'' \textit{IEEE Trans. Ind. Informat.}, vol. 18, no. 4, pp. 2571--2581, 2022. doi: \href{https://doi.org/10.1109/TII.2021.3103019}{10.1109/TII.2021.3103019}.

\bibitem{ref16}
X. Cheng, P.-S. Chen, Z. Wu, M. Cech, Z. Ying, and X. Hu, ``Automatic Detection of CFRP Subsurface Defects via Thermal Signals in Long Pulse and Lock-In Thermography,'' \textit{IEEE Trans. Instrum. Meas.}, vol. 72, pp. 1--10, 2023. doi: \href{https://doi.org/10.1109/TIM.2023.3277996}{10.1109/TIM.2023.3277996}.

\bibitem{ref17}
W. Hng Lim, S. Sfarra, T.-Y. Hsiao, and Y. Yao, ``Physics-Informed Neural Networks for Defect Detection and Thermal Diffusivity Evaluation in Carbon Fiber-Reinforced Polymer Using Pulsed Thermography,'' \textit{IEEE Trans. Instrum. Meas.}, vol. 74, pp. 1--10, 2025. doi: \href{https://doi.org/10.1109/TIM.2025.3551026}{10.1109/TIM.2025.3551026}.

\bibitem{ref18}
N. Yao, ``Subsurface Defect Detection in Carbon Fiber-Reinforced Plastics Using Infrared Polarization Imaging,'' \textit{IEEE Trans. Instrum. Meas.}, vol. 74, 2025. doi: \href{https://doi.org/10.1109/TIM.2025.3551026}{10.1109/TIM.2025.3551026}.

\bibitem{ref19}
H. Wang \textit{et al.}, ``Research on CFRP Small-Gap Defect Detection by Joint Scanning Laser Thermography Based on Restored Pseudo Heat Flux,'' \textit{IEEE Trans. Instrum. Meas.}, 2025. doi: \href{https://doi.org/10.1109/TIM.2025.3551026}{10.1109/TIM.2025.3551026}.

\bibitem{ref20}
L. Morelli, R. Marani, E. D'Accardi, D. Palumbo, U. Galietti, and T. D'Orazio, ``A convolution residual network for heating-invariant defect segmentation in composite materials inspected by lock-in thermography,'' \textit{IEEE Trans. Instrum. Meas.}, vol. 70, pp. 1--14, 2021. doi: \href{https://doi.org/10.1109/TIM.2021.3112345}{10.1109/TIM.2021.3112345}.

\bibitem{ref21}
P. Dhar, R. Amevorku, D. Amoateng-Mensah, and M. Sundaresan, ``Deep learning-based depth estimation of flat bottom holes in CFRP using active infrared thermography,'' \textit{SAMPE J.}, vol. 62, no. 2, pp. 38--44, 2026. doi: \href{https://doi.org/10.33599/SJ.v62no2.03}{10.33599/SJ.v62no2.03}.

\bibitem{ref22}
W. Zheng \textit{et al.}, ``A novel approach for one-step defect detection and depth estimation using sequenced thermal signal encoding,'' \textit{Nondestruct. Test. Eval.}, vol. 39, no. 8, pp. 2606--2621, 2024. doi: \href{https://doi.org/10.1088/10589759.2024.2304719}{10.1080/10589759.2024.2304719}.

\bibitem{ref23}
G. Zhou \textit{et al.}, ``Surface defect detection of CFRP materials based on infrared thermography and attention U-Net algorithm,'' \textit{Nondestruct. Test. Eval.}, vol. 39, no. 2, pp. 238--257, 2024. doi: \href{https://doi.org/10.1080/10589759.2023.2191954}{10.1080/10589759.2023.2191954}.

\bibitem{ref24}
M. Salah, N. Werghi, D. Svetinovic, and Y. Abdulrahman, ``Multi-modal attention networks for enhanced segmentation and depth estimation of subsurface defects in pulse thermography,'' \textit{arXiv preprint arXiv:2501.09994}, 2025. [Online]. Available: \url{https://arxiv.org/abs/2501.09994}

\bibitem{ref25}
P. Nooralishahi, B. M. Fisher, P. Collins, S. N. Givigi, and M. P. Brito, ``Drone-based non-destructive inspection of industrial sites: A review and case studies,'' \textit{Drones}, vol. 5, no. 4, p. 106, 2021. doi: \href{https://doi.org/10.3390/drones5040106}{10.3390/drones5040106}.

\bibitem{ref26}
K. Milionis, A. Eliades, V. Grigoriev, and M. J. Blanco, ``Unmanned Aerial Vehicles (UAVs) in the planning, operation and maintenance of concentrating solar thermal systems: A review,'' \textit{Solar Energy}, vol. 254, pp. 182--194, 2023. doi: \href{https://doi.org/10.1016/j.solener.2023.03.005}{10.1016/j.solener.2023.03.005}.

\bibitem{ref27}
Z. U. Abidin, H. Memon, and J. Ahmed, ``Three-Channel Thermographic Fusion for Architecture-Independent Defect Detection and Segmentation in Carbon Fibre Reinforced Polymer Composites Using Deep Learning,'' \textit{SSRN}, 2026. doi: \href{https://doi.org/10.2139/ssrn.7355934}{10.2139/ssrn.7355934}.

\bibitem{ref28}
L. Breiman, ``Random forests,'' \textit{Mach. Learn.}, vol. 45, no. 1, pp. 5--32, 2001. doi: \href{https://doi.org/10.1023/A:1010933404324}{10.1023/A:1010933404324}.

\bibitem{ref29}
T. Chen and C. Guestrin, ``XGBoost: A scalable tree boosting system,'' in \textit{Proc. 22nd ACM SIGKDD Int. Conf. Knowl. Discov. Data Min.}, San Francisco, CA, USA, 2016, pp. 785--794. doi: \href{https://doi.org/10.1145/2939672.2939785}{10.1145/2939672.2939785}.

\bibitem{ref30}
M. de Berg, O. Cheong, M. van Kreveld, and M. Overmars, \textit{Computational Geometry: Algorithms and Applications}, 3rd ed. Berlin, Germany: Springer, 2008. doi: \href{https://doi.org/10.1007/978-3-540-77974-2}{10.1007/978-3-540-77974-2}.

\bibitem{ref31}
J. Erazo-Aux, H. Loaiza-Correa, A. D. Restrepo-Gir{\'o}n, C. Ibarra-Castanedo, and X. Maldague, ``Thermal imaging dataset from composite material academic samples inspected by pulsed thermography,'' \textit{Data in Brief}, vol. 32, p. 106313, 2020. doi: \href{https://doi.org/10.1016/j.dib.2020.106313}{10.1016/j.dib.2020.106313}.

\bibitem{ref32}
J. Ahmed, B. Gao, and W. L. Woo, ``Wavelet-Integrated Alternating Sparse Dictionary Matrix Decomposition in Thermal Imaging CFRP Defect Detection,'' \textit{IEEE Trans. Ind. Informat.}, vol. 15, no. 7, pp. 4033--4043, 2019. doi: \href{https://doi.org/10.1109/TII.2018.2881341}{10.1109/TII.2018.2881341}.

\end{thebibliography}
\end{document}